\documentclass[runningheads]{llncs}

\usepackage{eccv}

\usepackage{eccvabbrv}

\usepackage{graphicx}
\usepackage{booktabs}

\usepackage[accsupp]{axessibility}  %
\usepackage{todonotes}
\usepackage{subcaption}
\usepackage{amsmath}
\usepackage{longtable}
\usetikzlibrary{arrows.meta,positioning,calc}

\usepackage[pagebackref,breaklinks,colorlinks,citecolor=eccvblue]{hyperref}

\usepackage{orcidlink}

\begin{document}

\title{Far from the Crowd: Scalable Self-Supervised Learning via Geographic Isolation} 

\titlerunning{Far from the Crowd}

\author{Daniele Rege Cambrin\inst{1}\orcidlink{0000-1111-2222-3333} \and
Francesco Rossi\inst{1}\orcidlink{1111-2222-3333-4444} \and
Mattia Varile\inst{1}\orcidlink{2222--3333-4444-5555}}

\authorrunning{D. Rege Cambrin et al.}

\institute{AIKO, Corso Stati Uniti 45, Turin, Italy
\email{\{name.surname\}@aikospace.com}\\
\url{https://www.aikospace.com/}}

\maketitle

\begin{abstract}
    Self-supervised pretraining on remote sensing imagery typically treats all samples as equally informative, despite large variability
    in geographic and visual structure. We propose a curriculum learning strategy for self-supervised Earth observation that ranks samples by
    geographic isolation, a label-free proxy derived entirely from geolocation metadata already present in geospatial datasets,  requiring no image
    decoding, no model feedback, and no manual annotation. Unlike visual complexity proxies, it scales as O(D log D) with
    dataset size D and is well-defined for both contrastive and reconstructive objectives.
    
    We integrate the proposed measure into MoCoV2 and MAE pretraining and evaluate across three downstream tasks from CopernicusBench
    (BigEarthNet, DFC-2020, LCZ). Our curriculum reaches baseline final-epoch performance using as few as 20\% of the training budget (MAE) and at most 40\% (MoCo) of the training budget,
    and improves final downstream performance by up to +5 mAP on BigEarthNet, with gains of 1–5 points across benchmarks, matching visual-complexity curricula while reducing pre-computation
    cost by more than 140× (4 s vs. 568 s on SSL4EO). A CKA and effective-rank analysis further reveals that curriculum-trained
    encoders develop higher-dimensional, more uniformly utilized embedding spaces throughout training.
\end{abstract}

\section{Introduction}

Self‑supervised learning has become the basis of modern visual representation learning, enabling models to learn rich features from large unlabeled datasets without manual annotations \cite{uelwer2025,Wang2022survey}. However, standard self‑supervised training treats all samples as equally important, ignoring that some images are inherently more difficult to model due to their visual or structural complexity. Curriculum learning seeks to address this by gradually exposing the model to easier examples first, thereby stabilizing optimization and improving generalization \cite{Wang2021,Soviany2022}. In the absence of explicit labels, defining a meaningful notion of difficulty for self‑supervised signals remains an open challenge.

In the labeled setting, difficulty is often grounded in semantic complexity (provided by the labels) or model uncertainty (defined by the model's supervised loss), but for self‑supervised learning these notions are not directly available (the labels are not present, and the loss does not represent the model uncertainty). Instead, one must rely on proxies that capture intrinsic properties of the input distribution.

The vast amount of historical Earth observation data is commonly used to build large-scale unlabeled pretraining corpora, particularly valuable given the significant expertise, time, and cost required for high-quality semantic annotation. \cite{Wang2022survey}. While they do not have associated labels, remote-sensed data (and geospatial data) have, in general, additional metadata: the geographic coordinates (\ie the location) \cite{ayush2021}.

In this work, we propose a scalable, model‑agnostic curriculum learning strategy for self‑supervised learning derived from the coordinates: geographic isolation. This approach relies on the observation that the most geographically isolated samples are also the most visually anomalous relative to the densely sampled regions of the dataset. Additionally, compared to visual complexity, it is more scalable for large datasets since its computational cost is $\mathcal{O}(D \log D + D\bar{k})$, where $\bar{k}$ is the mean neighborhood size, making it independent of image resolution and at least an order of magnitude cheaper than compression-based proxies at scale (a 142× speedup on SSL4EO). 

Experiments on using the well-established MoCoV2 \cite{chen2020} and MAE \cite{He2021} on SSL4EO \cite{Wang2022,blumenstiel2026} show that our curriculum improves convergence speed, representation quality, and downstream performance on BigEarthNet \cite{clasen2025}, DFC-2020 \cite{robinson2021}, LCZ \cite{zhu2020} compared to uniform sampling and visual complexity curriculum.  

In addition, we also provide a label-agnostic analysis of the learned feature space with central kernel alignment \cite{Kornblith2019} and effective rank \cite{olivier2007} to provide additional insight other than simply quantitative analysis on downstream tasks. 

Our contributions are:
\begin{itemize}
\item A scalable geographic isolation complexity measure that can be used in place of visual complexity;
\item A benchmark of the proposed curriculum learning on both well-established contrastive (MoCoV2) and reconstructive (MAE) approaches for SSL in Earth observation during the training evolution.
\item An embeddings space analysis over the training evolution to better understand the effects over the geometry composition. 
\end{itemize}

For reproducibility the code is released at \url{https://github.com/DarthReca/far-from-the-crowd}.

\section{Related Works}
This section covers the technique for self-supervised learning, curriculum learning, and their application to self-supervision. Finally, we cover XAI analysis for vision and earth observation.

\subsection{Self-Supervised Learning}
Self‑supervised learning methods learn representations from unlabeled data by designing pretext tasks that proxy downstream objectives. Contrastive learning frameworks such as MoCo \cite{He2019,chen2020}, SimCLR \cite{Chen2020simclr}, BYOL \cite{Grill2020}, DINO \cite{Caron2021,Oquab2023,Simeoni2025} maximize agreement between differently augmented views of the same image, while reconstruction‑based methods such as MAE \cite{He2021}, BEiT \cite{Bao2021} learn to reconstruct masked regions or transformations. In the geospatial domain, many unsupervised datasets such as CopernicusPretrain \cite{Wang2025}, EarthView \cite{Velzquez2025} and SSL4EO \cite{Stewart2023,Wang2022} and the relative models were created thanks to the large availability of data. These methods typically operate on large, heterogeneous collections of images but do not explicitly account for the varying difficulty of samples, instead sampling uniformly from the dataset.

Recent work has explored data‑aware sampling strategies in self‑supervised learning, such as curriculum‑based \cite{Soviany2022,Bengio2009,Hacohen2019} or uncertainty‑driven sampling \cite{Mindermann2022,swayamdipta2020}, but these often rely on trained models or semantic priors \cite{Cui2023,Sorscher2022} that are not available at initialization. In contrast, our approach defines difficulty without access to labels or model predictions, enabling earlier and more robust curriculum scheduling.

\subsection{Curriculum Learning}
Curriculum learning aims to improve training by gradually exposing the model to increasingly difficult examples. Early work in curriculum learning focused on ordered data, such as sorting captioned images by sentence length or complexity \cite{Bengio2009}, and has since been extended to automatic methods that estimate difficulty from model uncertainty or task‑specific metrics \cite{Kumar2010}.

In supervised settings, curriculum signals include predicted confidence, gradient norms, or losses, which can be used to select or reweight samples \cite{Soviany2022,Wang2021,Kumar2010}. However, these approaches are tied to the current state of the model and can be sensitive to initialization and early noise \cite{Wang2021,Kumar2010}. Several works have proposed more robust curriculum policies, such as self‑paced learning or dynamic difficulty scheduling, but they still require access to labels or model outputs \cite{Wang2021,weinshall2018,Higuchi2021,Soviany2022}.

In the self‑supervised setting, where neither labels or stable predictions are available early, designing a meaningful difficulty proxy without introducing model bias is particularly challenging. Our work sits at the intersection of these lines of research, providing a scalable, bias‑free curriculum signal grounded in information‑theoretic and spatial properties of the data.

\subsection{Curriculum Learning for Self-Supervision}
Curriculum learning has recently been adapted to self‑supervised representation learning, where the absence of labels demands alternative notions of sample difficulty. Srinidhi and Martel~\cite{Srinidhi2021} introduced a hardness-aware dynamic curriculum for contrastive pretraining leveraging annotator disagreement as a proxy for difficulty; however, their approach still depends on domain-specific metadata unavailable in generic large-scale datasets. u et al.~\cite{Lu2025} demonstrated that training on progressively cleaner data can induce noise robustness in self-supervised models without explicit denoising, suggesting that data ordering alone has a strong regularizing effect. Similarly, Joshi and Mirzasoleiman~\cite{Joshi2023} showed that examples most beneficial for contrastive SSL are structurally different from those useful for supervised learning, motivating the need for SSL-specific difficulty criteria.

Beyond individual sample selection, several works have investigated scheduling policies over training. Self-paced contrastive learning~\cite{Ge2020} introduces a learnable pace parameter that progressively incorporates harder negatives as training proceeds, though it requires the model to have already converged to a meaningful representation space before the curriculum can take effect. ReSim~\cite{Xiao2021} and subsequent region-based methods exploit local spatial structure to define patch-level difficulty, but assume access to object-level cues. In contrast, our method operates entirely at the image level, deriving a difficulty score from intrinsic properties of the data enabling curriculum scheduling from the very first epoch, without any model feedback or semantic annotations.

\section{Methodology}
In this section, we describe the complexity scores employed in this study along with the curriculum strategy employed.

\subsection{Complexity Scores}
In this section, we define how the complexity scores for the curriculum are defined given the lack of labels in SSL settings using both the common visual appearance and the geographical information.

\subsubsection{Visual Complexity}
A natural proxy for the visual difficulty of an unlabeled image is its information content. Intuitively, images with high spatial variability and diverse spectral patterns are harder to model than homogeneous ones. While entropy-based measures are commonly used, pixel-wise entropy ignores spatial structure, and embedding-based estimates introduce bias from pretrained representations.

We instead use lossless compression as a model-agnostic proxy for complexity, as it jointly captures spatial redundancy and spectral regularities. Given an image $i$ and a lossless compressor $c$ we define the compression ratio as:
$$
r = \frac {|i|}{|c(i)|}
$$

A high compression ratio indicates strong redundancy and low complexity (easy samples), whereas a low ratio reflects higher information density (hard samples). Unlike learned metrics, this proxy is independent of model inductive biases and requires no additional training or supervision and scales to hundreds of thousands of samples with negligible overhead on CPU.

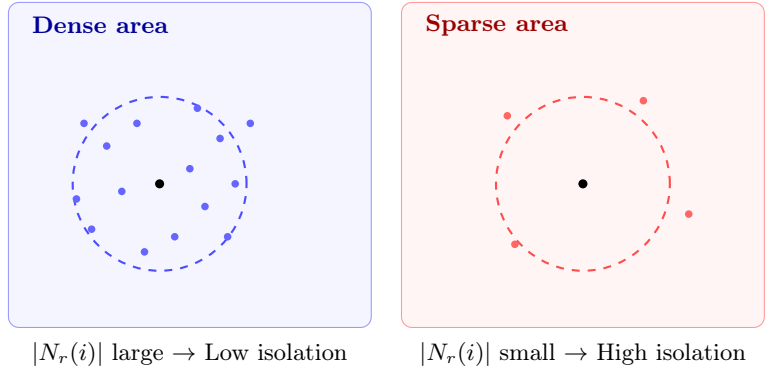
\begin{figure}[htb]
    \centering
    \begin{tikzpicture}[>=Latex, font=\sffamily]

\draw[rounded corners, fill=blue!4, draw=blue!40] (0,0.7) rectangle (4.8,5.0);
\node[anchor=west, font=\small\bfseries, text=blue!60!black] at (0.2,4.7) {Dense area};
\coordinate (qd) at (2.0,2.6);
\draw[thick, dashed, blue!70] (qd) circle (1.15);
\fill[black] (qd) circle (0.06);
\foreach \x/\y in {1.1/2.0,1.3/3.1,1.5/2.5,1.7/3.4,2.2/1.9,2.4/2.8,2.6/2.3,2.8/3.2,1.8/1.7,2.5/3.6,3.0/2.6,2.9/1.9,1.0/3.4,3.2/3.4,0.9/2.4}
  \fill[blue!60] (\x,\y) circle (0.05);
\node[align=center, font=\small] at (2.4,0.35) {$|N_r(i)|$ large $\to$ Low isolation};

\draw[rounded corners, fill=red!4, draw=red!40] (5.2,0.7) rectangle (10.0,5.0);
\node[anchor=west, font=\small\bfseries, text=red!60!black] at (5.4,4.7) {Sparse area};
\coordinate (qs) at (7.6,2.6);
\draw[thick, dashed, red!70] (qs) circle (1.15);
\fill[black] (qs) circle (0.06);
\foreach \x/\y in {6.6/3.5,8.4/3.7,9.0/2.2,6.7/1.8}
  \fill[red!60] (\x,\y) circle (0.05);
\node[align=center, font=\small] at (7.6,0.35) {$|N_r(i)|$ small $\to$ High isolation};

\end{tikzpicture}
    \caption{Given a fixed radius $r$, samples in densely covered areas have many neighbors and low isolation, whereas samples in sparsely covered areas have few neighbors and high isolation. This is aligned with the principle of spatial correlation.}
    \label{fig:geoiso_example}
\end{figure}

\subsubsection{Geographic Isolation Complexity}
Geospatial datasets provide implicit structural information through sample locations. According to Tobler’s first law of geography \cite{Tobler1970}:
\begin{quote}
    Everything is related to everything else, but near things are more related than distant things.
\end{quote}
so nearby samples tend to be more similar than distant ones. Consequently, geographically isolated samples are more likely to represent rare or diverse patterns (scenes that are underrepresented in the training distribution). Notably, this notion of rarity is orthogonal to visual complexity: empirically, geographic isolation and compression-ratio scores are rank-uncorrelated (Spearman $\approx 0$), suggesting they capture distinct data properties. We quantify this intuition using a local density estimate. For each sample $i$, we define its geographic isolation score as
$$
g = - \log{\bigg(\frac{|N_r(i)|}{\pi r^2} \bigg)}
$$
where $N_r(i)$ is the set of samples within radius $r$ of $i$, and $\pi r^2$  is the area of the neighborhood. The term inside the logarithm estimates the local sample density. Higher values of $g(i)$ correspond to lower-density regions (\ie more isolated and potentially more complex samples) as shown in \Cref{fig:geoiso_example}. The logarithm stabilizes the dynamic range and connects the measure to the notion of self-information \cite{Cover2005}.

To efficiently compute $N_r(i)$, we index geographic coordinates using a Ball Tree \cite{Omohundro2009}, enabling sub-linear query time and scalability to large datasets.

\begin{figure}[htb]
    \centering
    \begin{subfigure}{0.8\linewidth}
        \includegraphics[width=\linewidth]{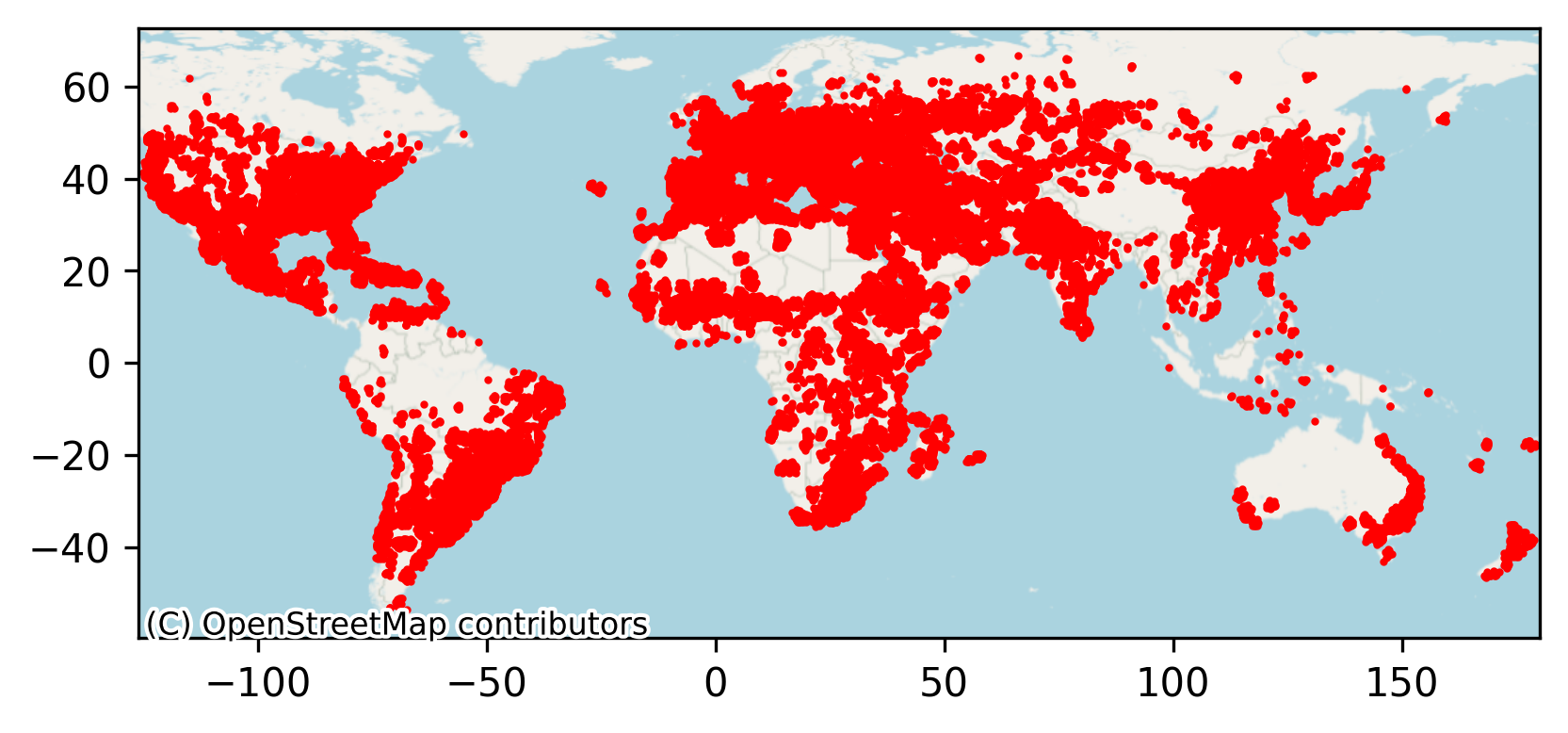}
        \caption{Whole Dataset}
    \end{subfigure}\\
    
    \begin{subfigure}{0.8\linewidth}
        \includegraphics[width=\linewidth]{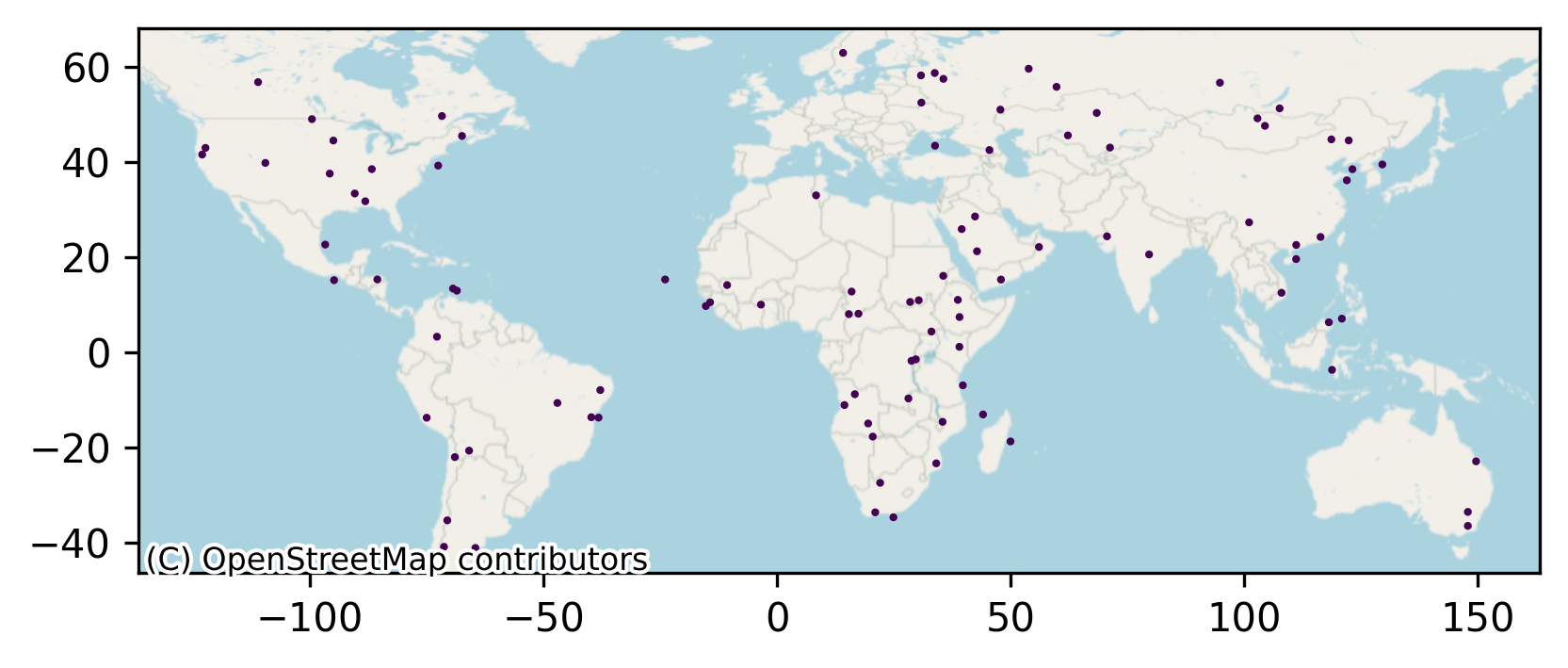}
        \caption{Top-100 Most Isolated Points}
    \end{subfigure}
    
    \caption{Geographical Distribution of Samples in the dataset and most geographically isolated points}
    \label{fig:geo_dist}
\end{figure}

It is possible to see the geographic distribution of the dataset and the top-100 most isolated points in \Cref{fig:geo_dist}.

\subsection{Curriculum Building}
Following a common definition in curriculum learning literature, we bin the difficulty scores into three, representing "easy," "medium," and "hard" samples \cite{Soviany2022} and then we apply a dynamic weight into the loss to each sample based on the difficulty tier \cite{Zhuang2024,Kumar2010,Bengio2009}. 

The reweighting scheme shared across MoCo and MAE is time-dependent. After assigning each sample to a difficulty tier $k$ (\ie easy, medium, hard), we modulate its contribution with an annealed weight $w_k(t)$.
$$
w_k(t)=\sigma\left(\alpha \frac{t-c_k}{T}\right)
$$
where $t$ is the current epoch, $T$ the total number of epochs, $c_k$ the activation center of tier $k$, $\alpha$ an annealing sharpness parameter. This schedule provides a soft curriculum as proposed in \cite{wang2019}: easy samples receive substantial weight earlier, while medium and hard samples are progressively emphasized as training advances and stabilize. 

In MoCo, these weights directly scale the instance-discrimination objective so that contrastive updates are curriculum-aware at the sample level. For MAE, we use the same tier-wise weighting on the per-sample reconstruction loss. In summary, instead of employing a mean, we aggregate using a sample-weighted mean. Additionally, this strategy permits the MoCo memory bank to have a high variety of samples from the whole dataset even if they do not contribute to the model update.

\section{Experiments}
In this section, we present the experimental settings and the experimental results. Additional settings are provided in supplementary.

\subsection{Experimental Settings}

\subsubsection{PreTraining}
For the SSL pretraining, we employed SSL4EOv1.1 Sentinel-2 subset \cite{blumenstiel2026} with around 250k locations around the globe. We trained a ResNet-50 \cite{kaiming2016} with MoCoV2 \cite{chen2020} and a ViT-Base with MAE \cite{He2021} to provide an overview of two different well-established training and architectural paradigms. We trained both networks with a similar wall-time budget ($\approx$12 h), which corresponds to 50 epochs for MoCoV2 and 100 epochs for MAE. The batch size was set to 128. The optimizer was SGD for MoCoV2 with a learning rate of 0.03 as in the original implementation and Adam with a learning rate of 1.5e-4 for MAE. The annealing value $\alpha$ was set to 8.0 after preliminary tests. The activation centers $c$ were set to 0\%, 20\%, and 40\% of training epochs. 

\subsubsection{Geographic Isolation} The geographic isolation depends on the ray $r$, which was set to 50 km. The choice is linked to the dataset construction: according to SSL4EOv1.1 \cite{blumenstiel2026} they sampled within a 50 km radius of the world’s 10,000 most populated cities, ensuring near-global coverage while focusing on urban areas but also including ocean patches and rural areas. Additionally, if we fit a BallTree with the dataset coordinates and we query each point for the 50 km radius, we get more than one neighbor for 99.8\% of the samples (vs. 97\% for 20 km), confirming the sampling strategy of the paper.

\subsubsection{Linear Probing}
We selected three tasks from CopernicusBench from the Sentinel-2 subset \cite{Yi2025} evaluated according to the original metrics: two level 2 tasks (\ie BigEarthNet-S2 and DFC-2020-S2) and one level 3 task (\ie LCZ-S2) to test the goodness of the representation for different difficulties. We tested all the tasks in linear probing to better evaluate the quality of the learned representations, adding a linear layer for classification and a convolutional layer for segmentation. 

\subsubsection{Evaluation Metrics}
We evaluate the performance according to the original metrics of CopernicusBench \cite{Yi2025}: multilabel classification (ML Cls) with 19 classes of BigEarthNet-S2 using mean average precision (mAP), multiclass semantic segmentation (MC Seg) with 10 classes of DFC-2020-S2 using micro Intersection-over-Union (mIoU), and multiclass classification (MC Cls) with 17 classes of LCZ-S2 using overall accuracy (OA).
We report the linear probing results across three different seeds, and we performed Bayesian tests \cite{benavoli2017} with a rope of 0.5 to account for significant differences with small samples (\ie 3 seeds). You can find all the tests in the supplementary materials.

\subsection{Experimental Results}
We evaluate representations at five equally spaced checkpoints covering 20\%, 40\%, 60\%, 80\%, and 100\% of the total training budget (epochs 10, 20, 30, 40, 50 for MoCoV2; 20, 40, 60, 80, 100 for MAE) to assess not only whether final representation quality improves, but also whether convergence is accelerated. The tabulated results can be found in supplementary material.

\begin{figure}[htb]
    \centering
    \begin{subfigure}{\linewidth}
        \includegraphics[width=\linewidth]{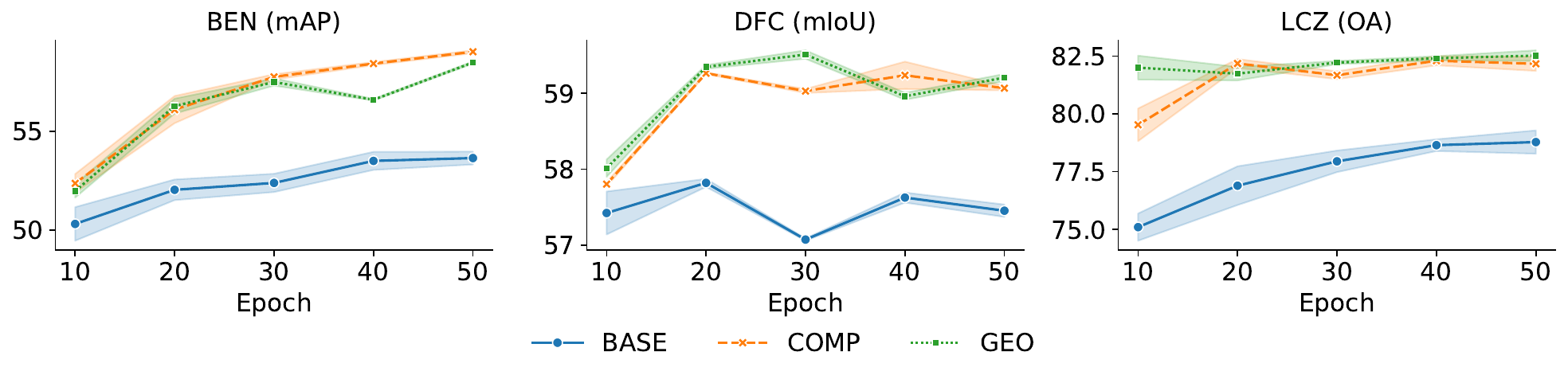}
        \caption{MoCoV2 performance by epoch}
        \label{fig:moco_trend}
    \end{subfigure} \\

    \vspace{5mm}
    
    \begin{subfigure}{\linewidth}
        \includegraphics[width=\linewidth]{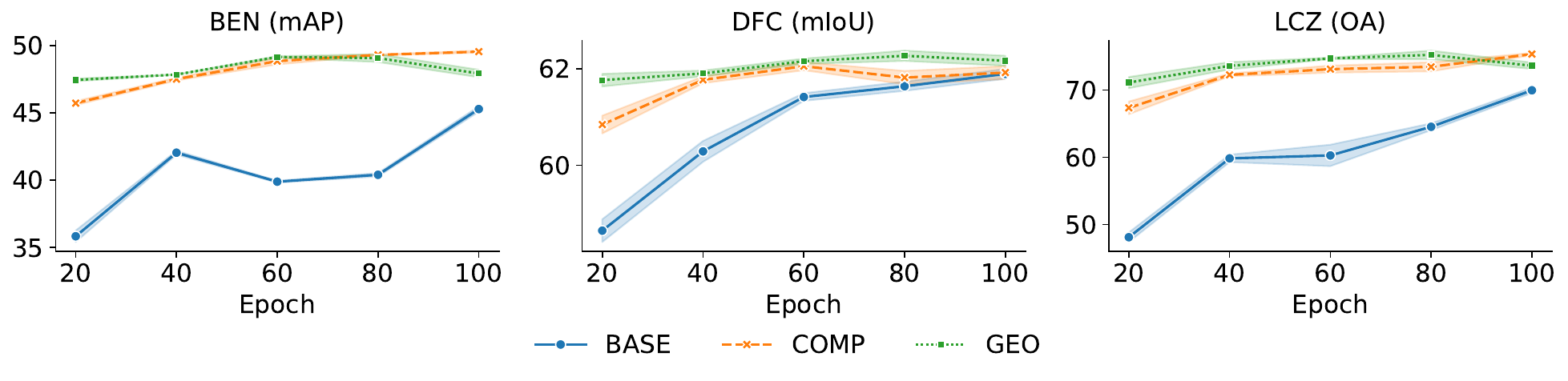}
        \caption{MAE performance by epoch}
        \label{fig:mae_trend}
    \end{subfigure}

    \caption{Downstream linear probing performance along SSL pretraining. Results are reported for MoCoV2 and MAE using the baseline training (BASE), the compression-ratio curriculum (COMP), and the geographic-isolation curriculum (GEO). The bands report the standard deviation.}
    \label{fig:main_trends}
\end{figure}

As shown in \Cref{fig:main_trends}, curriculum-based pretraining consistently improves the quality of the learned representations with respect to the non-curriculum baseline across both SSL paradigms. This effect is particularly evident during the early stages of pretraining, where both curricula reach substantially higher downstream performance than the baseline after only a fraction of the total training budget. This suggests that the proposed reweighting strategy does not only improve final representations but also accelerates the emergence of transferable features.

For MoCoV2, both curriculum variants outperform the baseline on all downstream tasks during the whole training, also in early epochs. The Bayesian tests confirm this trend with probabilities higher than 95\%, except for epoch 10 in DFC. On DFC-2020, the improvements are more moderate in absolute terms, likely due to saturation of the linear separability, but both curriculum strategies still provide a consistent advantage over the baseline. For MAE, the gains induced by curriculum learning are even more pronounced. The baseline shows a much slower convergence, especially on BigEarthNet-S2 and LCZ-S2, where the performance remains substantially below the curriculum variants for most of training. For DFC-2020, the baseline progressively closes part of the gap (as also indicated by the Bayesian test with probability 99\%), but the curriculum-based methods remain consistently superior or comparable across all checkpoints. These results indicate that MAE benefits strongly from presenting easier samples with higher weight in the initial stages, likely because reconstruction-based pretraining is particularly sensitive to the visual complexity of the input distribution.

Comparing the two complexity scores, neither criterion dominates uniformly; Bayesian equivalence probabilities range from 70\% to 90\% across checkpoints. Critically, this parity is achieved at dramatically different costs, making geographic isolation the preferred choice when compute is constrained. This suggests the downstream benefit is not proxy-specific but reflects a general principle: any curriculum that breaks uniform sampling accelerates representation learning in the EO pretraining setting.

\subsection{Computational Cost}
Although dynamically adapting the curriculum during training is attractive in principle, pre-computing a fixed ordering provides a less adaptive but far more scalable solution — particularly suited to large datasets, repeated runs, and long training schedules.

\paragraph{Visual Complexity} When using a compression algorithm, we can assume the worst case as a linear cost to the number of pixels $P$ in an image, so for a dataset of size, $D$ the cost is $O(D\cdot N)$. However, it is important to note that the process can be parallelized.

\paragraph{Geographic Isolation Complexity} For geographic isolation via a BallTree, the one-time construction cost is $\mathcal{O}(D \log D)$ and each radius query costs $\mathcal{O}(\log D + k)$, where $k = |N_r(i)|$. Summed over the full dataset, the total cost is $\mathcal{O}(D \log D + D\bar{k})$, which reduces to $\mathcal{O}(D \log D)$ when $\bar{k} \ll D$, as is typical for geographically distributed data

When using a BallTree to compute the geographic isolation, we incur (for a 2D representation) into a building cost of $O(D\log(D))$, while each query cost $O(\log(D) + k)$, where $k = |N_r(i)| << D$, if the dataset is geographically distributed (the worst case is $k = D$ if all the points are returned). So for the dataset $D$, the whole cost to compute the geographic isolation is $O(D \log D + D \overline{k})$, where $\overline{k}$ is the mean number of neighbors.

\paragraph{Self-Paced Learning} The usage of self-paced learning is generally costly and not suited for this setting, since readapting the difficulty each epoch costs $O(T \cdot D \cdot F_{model})$, where $T$ is the number of epochs and $F_{model}$ is the forward pass cost of the model.

Our one-shot solution can be cheaply computed without slowing the training independently of the model or the training paradigm without relying on labels. Compared to compression, it is not dependent on the image size but only on the number of samples and can be easily computed with few resources. In practice, on the full SSL4EO dataset: geographic isolation (BallTree) takes 4 s; compression ratio takes 568 s (142× slower); a single ResNet-50 forward pass over the dataset takes 379 s at batch size 128 and 34,962 s at batch size 1, two to four orders of magnitude slower than geographic isolation, and is repeated every epoch for self-paced learning.

\subsection{Additional Experimental Results}

\subsubsection{Performance over Masking Ratio Curriculum}
Following previous approaches \cite{Naskar2025,Yoon2025}, we also evaluate how the proposed approach compares to gradually increasing the masking ratio of MAE as another scalable solution for SSL (which we name CLM hereafter).

\begin{figure}
    \centering
    \includegraphics[width=\linewidth]{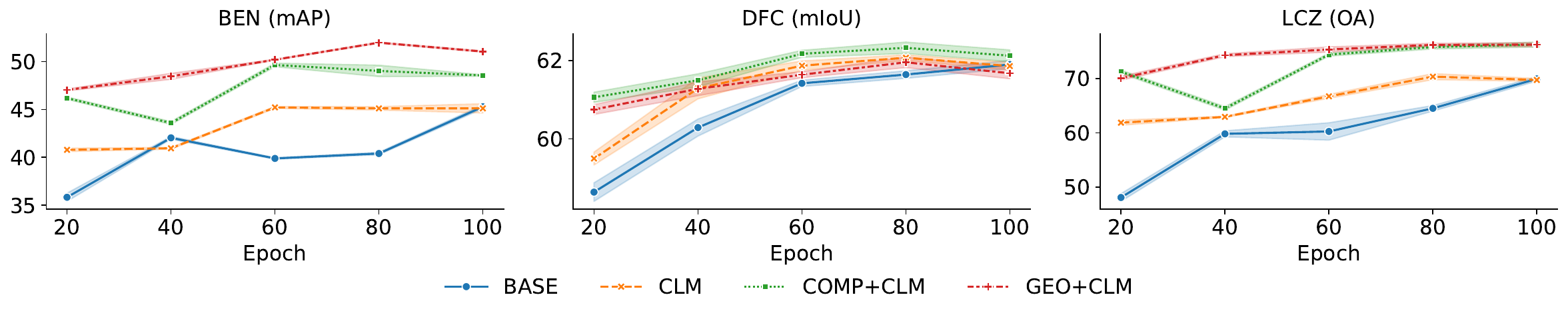}
    \caption{Downstream performance across training epochs for the baseline MAE (BASE), masking ratio curriculum (CLM), and masking ratio curriculum combined with compression-based (COMP+CLM) and geographic isolation-based (GEO+CLM) sample ordering. Shaded areas denote standard deviations.}    
    \label{fig:clm_performance}
\end{figure}

As shown in \Cref{fig:clm_performance}, the masking-ratio curriculum already improves over the fixed MAE baseline in most settings. Nevertheless, combining CLM with the proposed sample-level curricula further improves performance, particularly in the early and intermediate stages of pretraining. This indicates that task-level difficulty (masking ratio) and data-level difficulty (sample weighting) are orthogonal axes of curriculum design: each contributes an independent signal, and combining them is strictly better than either alone, particularly in early pretraining.

The Bayesian tests support this observation. On BigEarthNet-S2, both compression and geographic isolation improve over CLM with probabilities above $98\%$. On DFC-2020, the combined curricula are better at epoch 20, with probabilities above $97\%$, while they become statistically equivalent to CLM at later checkpoints, with equivalence probabilities between $85\%$ and $99\%$, suggesting partial saturation of the downstream task. On LCZ-S2, both proposed combinations remain better than CLM with probabilities above $97\%$. Comparing the two sample-level proxies, GEO+CLM is generally better than or equivalent to COMP+CLM, with probabilities ranging from $80\%$ to $99\%$.

\subsubsection{Additional Metrics Analysis}
\begin{figure}[htb]
    \centering

    \begin{subfigure}{\linewidth}
        \centering
        \includegraphics[width=\linewidth]{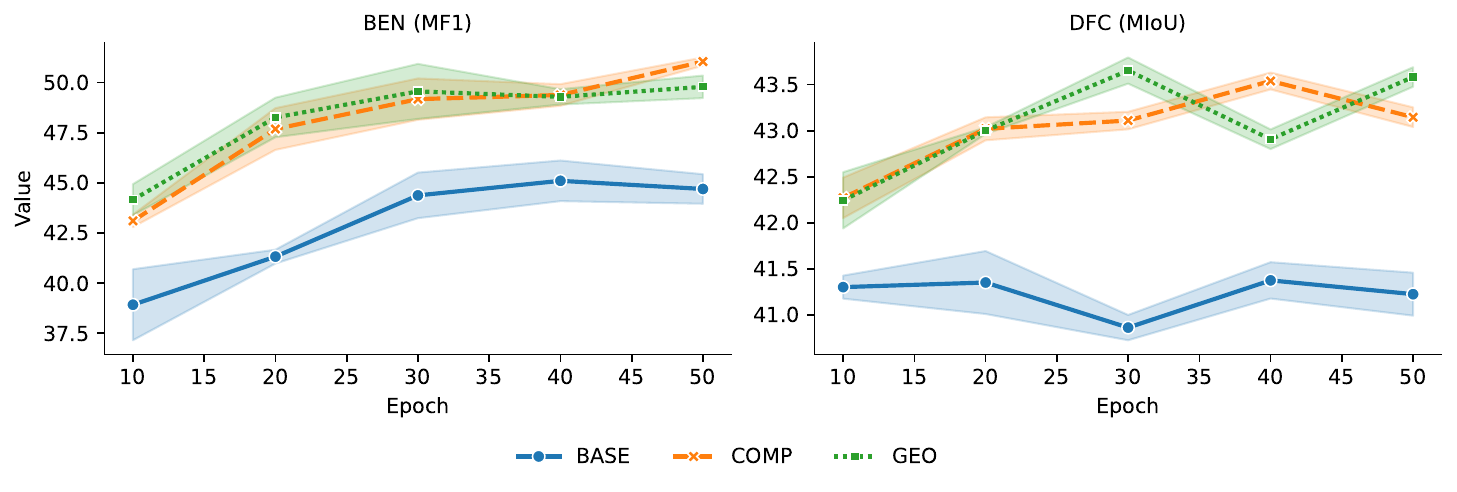}
    \caption{Macro F1 (BEN) and Macro IoU (DFC) across training epochs for MoCo-based pretraining. }      \label{fig:macro_moco_trend}
    \end{subfigure}\\

    \begin{subfigure}{\linewidth}
        \centering
        \includegraphics[width=\linewidth]{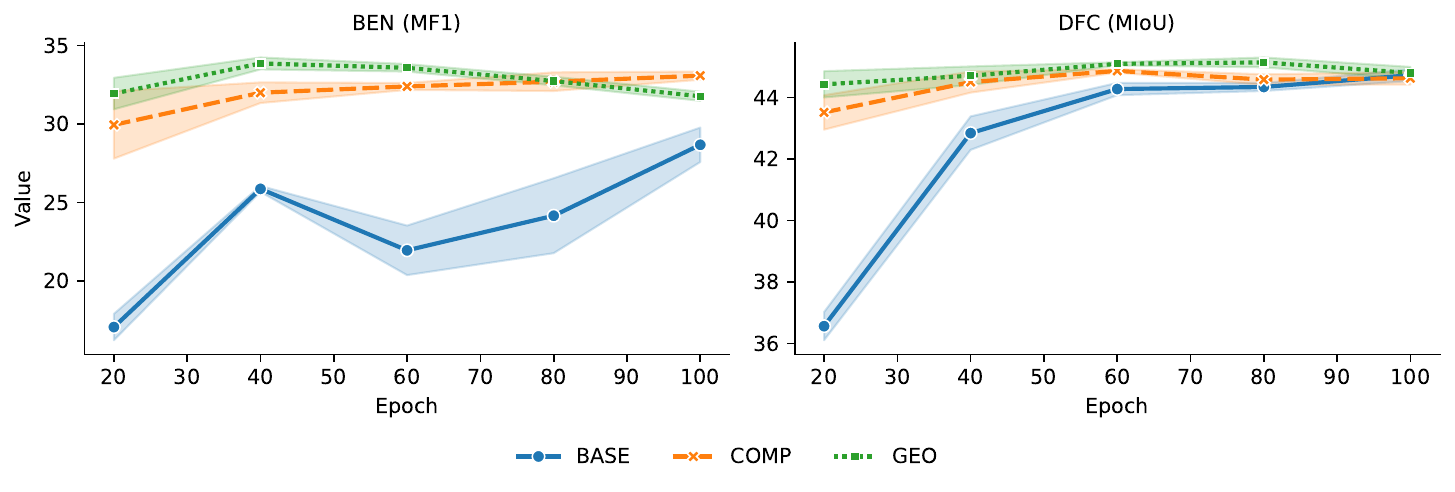}
    \caption{Macro F1 (BEN) and Macro IoU (DFC) across training epochs for MAE-based pretraining}        \label{fig:macro_mae_trend}
    \end{subfigure}

    \caption{Macro F1 on BEN and Macro IoU on DFC across pretraining epochs for MoCo and MAE.}
    \label{fig:macro_trends}
    
\end{figure}

LCZ already balances the classes by construction, so overall accuracy is meaningful enough, while for BigEarthNet and DFC, such balance is not guaranteed by the dataset. Along with the standard metrics for the tasks proposed in CopernicusBench, we report the macro F1 scores, which let better understand the performance for rare classes as shown in \Cref{fig:macro_trends}.

On BigEarthNet, both curriculum strategies substantially improve macro F1 over the baseline for both SSL paradigms, indicating that the learned representations are not only improving dominant classes but also better supporting rarer or under-represented land-cover categories (this is also confirmed by a Bayesian test with probability > 95\%). 
For DFC, using MAE, the curriculum variants provide a clear advantage at the early checkpoints (with probability of $\approx 98\%$), after which the baseline progressively closes the gap and all methods converge to similar MacroIoU values (with probability of equality $\approx 98\%$ for the last two checkpoints). This suggests that, on DFC, curriculum learning mainly accelerates convergence, while the final representations become comparable once pretraining is sufficiently long. The two variants of curriculum behave similarly, and there is no clear winner (the Bayesian test confirms this finding, suggesting a probability higher than 97\% of equality for most of the checkpoints).

Overall, the class-balanced analysis reinforces the main experimental findings. The proposed curricula improve convergence speed and produce representations that are more robust to class imbalance, which is particularly relevant in remote-sensing datasets where rare semantic categories are common.

\section{Embedding Space Analysis}
To better understand how the embeddings space is built during the epochs, we analyze the embeddings space (of the last layer) of the test set of BigEarthNet (with around 6k samples), since it is the most semantically varied dataset, using a label-agnostic analysis, to provide a more general understanding of the model behavior. 

\subsection{Central Kernel Alignment}
\begin{figure}[htb]
    \centering
    \begin{subfigure}{0.8\linewidth}
        \includegraphics[width=\linewidth]{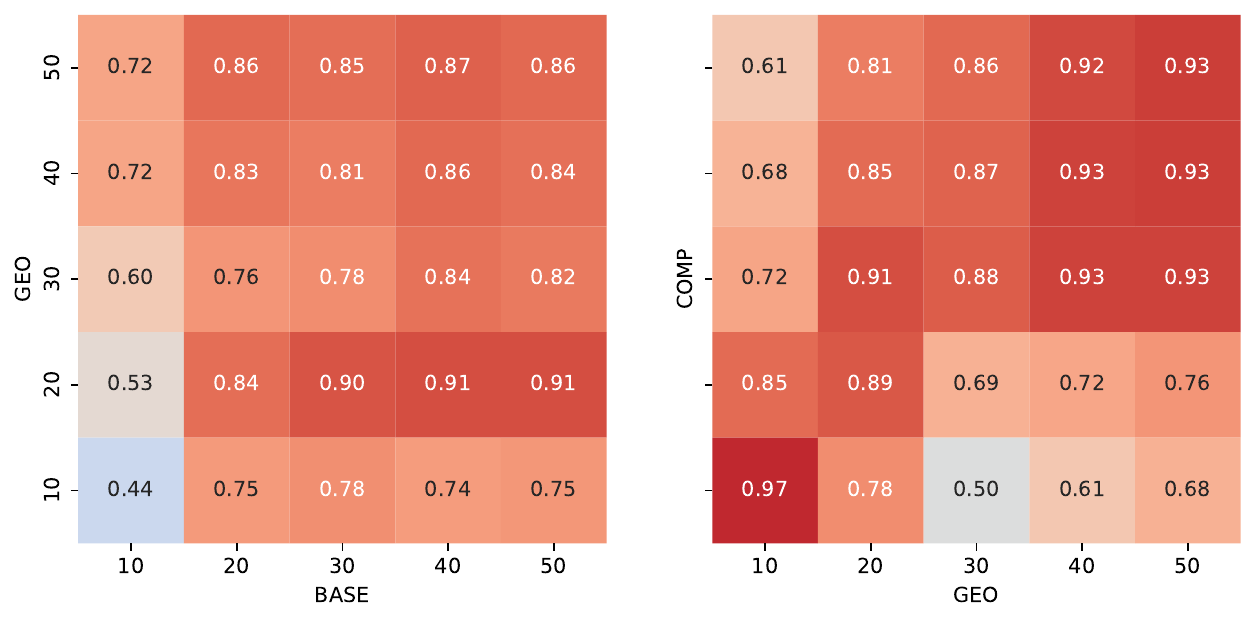}
        \caption{MoCo}
    \end{subfigure} \\
    \begin{subfigure}{0.8\linewidth}
        \includegraphics[width=\linewidth]{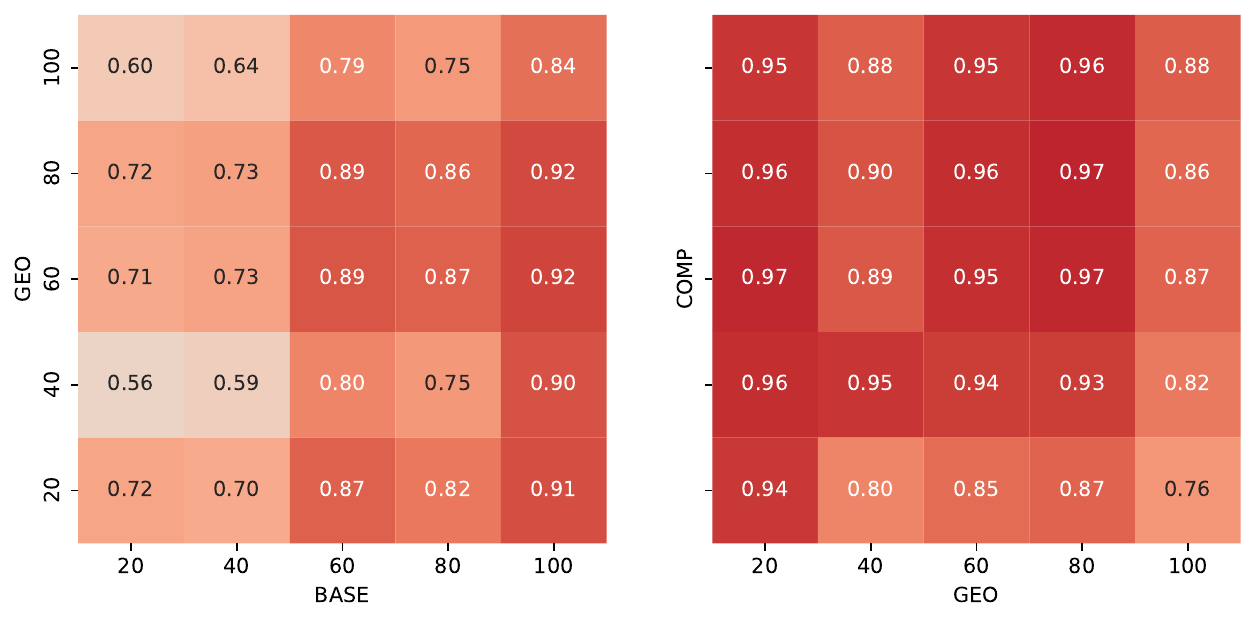}
        \caption{MAE}
    \end{subfigure}
    \caption{Central Kernel Alignment (CKA) between encoder representations at different pretraining epochs, computed on the
BigEarthNet test set.}
    \label{fig:cka_heatmaps}
\end{figure}

First, we analyze using linear Central Kernel Alignment \cite{Kornblith2019} to quantify how representational geometry evolves across training and to determine whether curricula accelerate or alter the trajectory of representational convergence \cite{Shekhar2023} in \Cref{fig:cka_heatmaps}.

For MoCoV2, the BASE-vs-GEO matrix shows the geographic-isolation curriculum alters the early-stage representation geometry, but its effect on the final representation becomes more moderate, so the baseline partially "catches up" in terms of which features it learns. The GEO-vs-COMP matrix, in contrast, suggests that they converge to similar representations as pretraining progresses, providing an explanation for the empirical observation that the two curricula are often statistically equivalent on downstream tasks.

For MAE, both matrices are uniformly higher, indicating that MAE-based encoders, regardless of the curriculum, end up learning more similar representations than MoCo-based ones. This also suggests that generally employing a curriculum-induced training diversity enhances representation quality. The BASE-vs-GEO similarity mirrors the MoCo trend, while the GEO-vs-COMP matrix is remarkably saturated, with the lowest values appearing at the diagonal extremes (epochs 20 and 100). This suggests that, for masked reconstruction, the two curricula act through a similar mechanism (\ie modulating the visual complexity distribution seen by the encoder) and that this mechanism dominates over the choice of proxy.

\subsection{Effective Rank}
\begin{figure}[htb]
    \centering
    \begin{subfigure}{0.48\linewidth}
        \includegraphics[width=\linewidth]{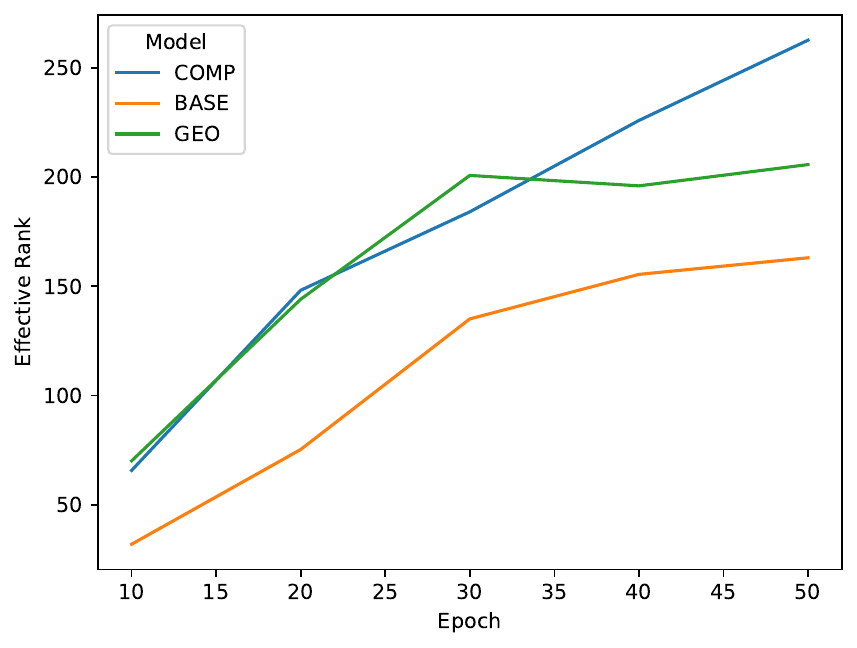}
        \caption{MoCo}
    \end{subfigure}
    \hfill
    \begin{subfigure}{0.48\linewidth}
        \includegraphics[width=\linewidth]{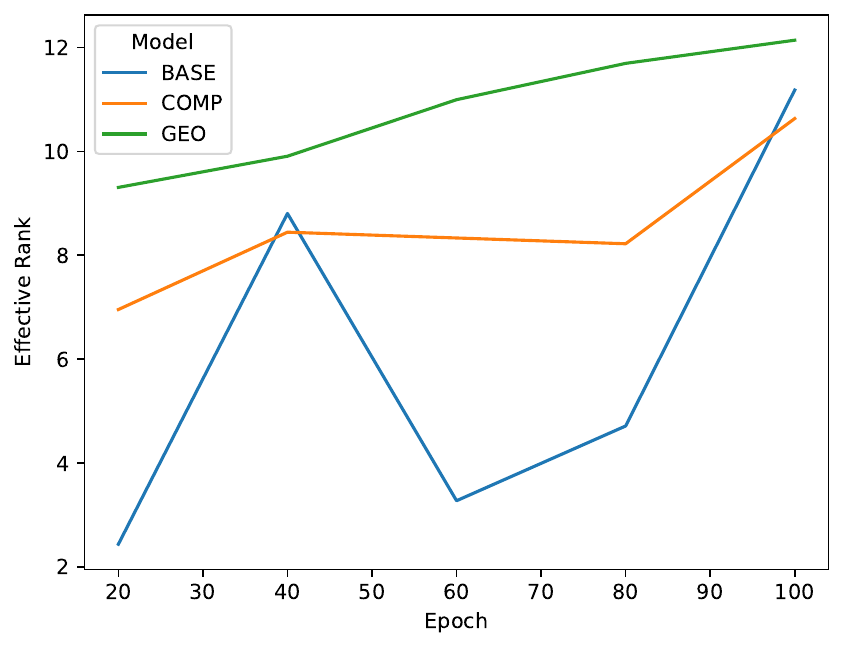}
        \caption{MAE}
    \end{subfigure}
    \caption{Effective rank of the encoder representations on the
BigEarthNet test set across pretraining epochs under the baseline (BASE), the compression-ratio
curriculum (COMP), and the geographic-isolation curriculum (GEO).}
    \label{fig:effective_rank}
\end{figure}

Second, we analyze the effective rank \cite{olivier2007} to understand how much the embeddings space is exploited \cite{Jing2021} in \Cref{fig:effective_rank}.

For MoCoV2, both compression and geo-isolation reach effective ranks well above BASE throughout training. Interestingly, the two curricula exhibit different growth dynamics: COMP grows almost monotonically across epochs and attains the highest final effective rank, whereas GEO plateaus. This suggests that the compression-based proxy yields a more gradual, sustained diversification of the representation, while the geographic-isolation proxy provides an early, strong diversification signal that the model quickly absorbs. The baseline, in contrast, grows much more slowly and never catches up within the same training budget, indicating that, without an explicit curriculum, instance discrimination tends to compress the embedding onto a narrower subspace.

For MAE, the effect is qualitatively similar but markedly more unstable. The baseline exhibits large epoch-to-epoch fluctuations in effective rank (a well-known signature of the instability known to affect masked-image modeling \cite{Zhang2022,He2021,Shekhar2023} when reconstruction targets are uniformly complex). Both COMP and GEO suppress these oscillations and maintain consistently higher ranks, with GEO providing the smoothest trajectory, suggesting it acts as a distributional regularizer on the reconstruction difficulty seen during early training.

\section{Conclusion}
Geographic isolation provides an effective curriculum signal for self-supervised Earth observation: without any model feedback, label, or image decoding, a single BallTree construction and query over coordinates (costing just 4 s for SSL4EO, versus 568 s for compression-based alternatives) suffices to substantially accelerate representation learning and improve downstream performance.

Experiments with MoCoV2 and MAE on SSL4EO-v1.1, evaluated across three downstream tasks from CopernicusBench, show that both curriculum variants consistently accelerate convergence and improve final downstream performance over uniform sampling. The gains are most pronounced in the early pretraining regime and for MAE, where reconstruction objectives are particularly sensitive to the complexity distribution of the training data. The embedding space analysis reinforces these findings: curriculum-trained encoders occupy a higher-dimensional, more isotropic representation space throughout training, and CKA analysis reveals that both proxies converge to geometrically similar solutions once pretraining is sufficiently long. This robustness to proxy choice is a practically useful finding: geographic isolation and compression ratio are largely interchangeable in terms of representational outcome, yet the former is dramatically cheaper to compute. This also suggests that applying a curriculum strategy is relevant to improve the representational quality regardless of the proxy.

Future work could extend geographic isolation to multi-modal geospatial datasets (optical, SAR, and elevation); to temporal sequences where both spatial and temporal isolation are meaningful; and to continual learning settings where new acquisitions arrive incrementally and re-ranking must remain cheap.

\bibliographystyle{splncs04}
\bibliography{main}

\appendix

\section{Additional Experimental Settings}
All run were performed on a single NVIDIA A6000.

\subsection{Compression Ratio} Since we are simply interested in the redundancy, we compress the image data with zlib \footnote{\url{https://docs.python.org/3/library/zlib.html}} with a compression level of 1 for minimal computational burden.

\subsection{Linear Probing}
The pretrained backbones were frozen. The batch size was set to 128. We used the AdamW optimizer with reduce learning rate on plateau for 30 epochs and a learning rate of 1e-3. We employ binary cross-entropy loss for BigEarthNet and cross-entropy loss for LCZ. The bands that are not included in the downstream tasks are zeroed (the Sentinel-2 NODATA value).

\section{Full Bayesian Comparison}
The results for the Bayesian tests over the main metrics are reported in \Cref{tab:bay_test_main}, while for macro metrics in \Cref{tab:bay_test_macro} for both MoCov2 and MAE during the training epochs. The results of the Bayesian tests over the curriculum masking variants are reported in \Cref{tab:bay_test_clm}.

\section{Tabulated Results}
We report the results for each seed (\ie 123, 0, 42) for both MoCo and MAE variants for the main metrics in \Cref{tab:main_results} and for the macro metrics in \Cref{tab:macro_results}. In \Cref{tab:clm_results}, we report the results for curriculum masking (CLM) variants across the same seeds.

\begin{longtable}{@{}ccc|ccc|ccc@{}}
    \caption{Bayesian Test Probability on main metrics with Rope of 0.5}
    \label{tab:bay_test_main} \\
Dataset & Model        & Epoch (\%) & \multicolumn{3}{c|}{MoCo}    & \multicolumn{3}{c}{MAE}      \\
        &              &            & P(left) & P(right) & P(rope) & P(left) & P(right) & P(rope) \\ \midrule\endhead

    \bottomrule
    \multicolumn{2}{r}{{Continued on next page}} \\
    \endfoot
    
    \endlastfoot
        
BEN     & GEO vs BASE  & 20\%       & 92.99   & 2.38     & 4.63    & 99.96   & 0.03     & 0.01    \\
BEN     & GEO vs BASE  & 40\%       & 97.74   & 1.45     & 0.82    & 99.97   & 0.02     & 0.01    \\
BEN     & GEO vs BASE  & 60\%       & 99.43   & 0.39     & 0.18    & 100.00  & 0.00     & 0.00    \\
BEN     & GEO vs BASE  & 80\%       & 98.87   & 0.60     & 0.53    & 99.89   & 0.09     & 0.02    \\
BEN     & GEO vs BASE  & 100\%      & 99.82   & 0.12     & 0.06    & 98.60   & 0.67     & 0.74    \\
BEN     & COMP vs BASE & 20\%       & 96.35   & 1.45     & 2.19    & 99.96   & 0.03     & 0.01    \\
BEN     & COMP vs BASE & 40\%       & 95.87   & 2.64     & 1.48    & 100.00  & 0.00     & 0.00    \\
BEN     & COMP vs BASE & 60\%       & 99.67   & 0.23     & 0.10    & 99.94   & 0.05     & 0.01    \\
BEN     & COMP vs BASE & 80\%       & 99.69   & 0.21     & 0.10    & 99.99   & 0.01     & 0.00    \\
BEN     & COMP vs BASE & 100\%      & 99.88   & 0.08     & 0.04    & 99.86   & 0.08     & 0.05    \\
BEN     & GEO vs COMP  & 20\%       & 1.22    & 25.94    & 72.84   & 99.55   & 0.14     & 0.32    \\
BEN     & GEO vs COMP  & 40\%       & 20.06   & 7.87     & 72.07   & 9.37    & 0.48     & 90.15   \\
BEN     & GEO vs COMP  & 60\%       & 0.87    & 7.11     & 92.02   & 28.47   & 5.32     & 66.21   \\
BEN     & GEO vs COMP  & 80\%       & 0.05    & 99.85    & 0.10    & 5.73    & 21.86    & 72.41   \\
BEN     & GEO vs COMP  & 100\%      & 0.02    & 91.03    & 8.94    & 1.13    & 96.28    & 2.59    \\
DFC     & GEO vs BASE  & 20\%       & 58.85   & 4.56     & 36.59   & 99.77   & 0.12     & 0.11    \\
DFC     & GEO vs BASE  & 40\%       & 99.79   & 0.05     & 0.15    & 99.12   & 0.25     & 0.63    \\
DFC     & GEO vs BASE  & 60\%       & 99.98   & 0.01     & 0.01    & 96.40   & 0.15     & 3.45    \\
DFC     & GEO vs BASE  & 80\%       & 99.87   & 0.03     & 0.11    & 97.74   & 0.03     & 2.23    \\
DFC     & GEO vs BASE  & 100\%      & 99.88   & 0.04     & 0.08    & 0.19    & 0.02     & 99.80   \\
DFC     & COMP vs BASE & 20\%       & 35.72   & 4.53     & 59.74   & 99.88   & 0.05     & 0.07    \\
DFC     & COMP vs BASE & 40\%       & 99.84   & 0.04     & 0.12    & 99.09   & 0.23     & 0.68    \\
DFC     & COMP vs BASE & 60\%       & 100.00  & 0.00     & 0.00    & 83.08   & 0.48     & 16.44   \\
DFC     & COMP vs BASE & 80\%       & 98.45   & 0.44     & 1.11    & 0.76    & 0.17     & 99.07   \\
DFC     & COMP vs BASE & 100\%      & 99.81   & 0.05     & 0.14    & 0.55    & 0.45     & 99.00   \\
DFC     & GEO vs COMP  & 20\%       & 3.71    & 0.68     & 95.61   & 97.42   & 0.25     & 2.34    \\
DFC     & GEO vs COMP  & 40\%       & 0.06    & 0.03     & 99.92   & 0.20    & 0.07     & 99.74   \\
DFC     & GEO vs COMP  & 60\%       & 29.59   & 0.05     & 70.36   & 0.58    & 0.25     & 99.17   \\
DFC     & GEO vs COMP  & 80\%       & 1.93    & 14.56    & 83.51   & 16.67   & 0.07     & 83.26   \\
DFC     & GEO vs COMP  & 100\%      & 0.13    & 0.04     & 99.83   & 1.09    & 0.13     & 98.78   \\
LCZ     & GEO vs BASE  & 20\%       & 99.79   & 0.15     & 0.05    & 99.96   & 0.03     & 0.00    \\
LCZ     & GEO vs BASE  & 40\%       & 98.04   & 1.32     & 0.64    & 99.85   & 0.13     & 0.02    \\
LCZ     & GEO vs BASE  & 60\%       & 99.17   & 0.52     & 0.31    & 99.38   & 0.54     & 0.08    \\
LCZ     & GEO vs BASE  & 80\%       & 99.53   & 0.27     & 0.19    & 99.94   & 0.05     & 0.01    \\
LCZ     & GEO vs BASE  & 100\%      & 98.10   & 1.13     & 0.76    & 99.97   & 0.02     & 0.01    \\
LCZ     & COMP vs BASE & 20\%       & 99.95   & 0.03     & 0.02    & 99.78   & 0.20     & 0.02    \\
LCZ     & COMP vs BASE & 40\%       & 98.60   & 0.97     & 0.43    & 99.94   & 0.05     & 0.01    \\
LCZ     & COMP vs BASE & 60\%       & 98.66   & 0.79     & 0.55    & 99.34   & 0.57     & 0.09    \\
LCZ     & COMP vs BASE & 80\%       & 99.60   & 0.23     & 0.17    & 99.85   & 0.12     & 0.03    \\
LCZ     & COMP vs BASE & 100\%      & 99.81   & 0.11     & 0.09    & 99.88   & 0.09     & 0.04    \\
LCZ     & GEO vs COMP  & 20\%       & 97.87   & 0.97     & 1.16    & 97.27   & 1.65     & 1.07    \\
LCZ     & GEO vs COMP  & 40\%       & 0.52    & 29.90    & 69.58   & 92.06   & 2.04     & 5.90    \\
LCZ     & GEO vs COMP  & 60\%       & 62.46   & 0.73     & 36.81   & 92.01   & 2.63     & 5.36    \\
LCZ     & GEO vs COMP  & 80\%       & 9.48    & 4.73     & 85.79   & 99.58   & 0.13     & 0.29    \\
LCZ     & GEO vs COMP  & 100\%      & 39.30   & 10.66    & 50.04   & 0.96    & 97.04    & 2.00    \\ \bottomrule
\end{longtable}

\begin{longtable}{@{}ccc|ccc@{}}
    \caption{Bayesian Test Probability on main metrics for Curriculum masking ratio (CLM) with Rope of 0.5}
    \label{tab:bay_test_clm} \\
    
    \toprule
    Dataset & Model & Epoch & P(left) & P(right) & P(rope) \\
    \midrule\endhead
    
    \bottomrule
    \multicolumn{2}{r}{{Continued on next page}} \\
    \endfoot
    
    \endlastfoot
        
    BEN & GEO+CLM vs CLM & 20\% & 99.91 & 0.06 & 0.02 \\
    BEN & GEO+CLM vs CLM & 40\% & 99.87 & 0.10 & 0.03 \\
    BEN & GEO+CLM vs CLM & 60\% & 99.99 & 0.01 & 0.00 \\
    BEN & GEO+CLM vs CLM & 80\% & 99.98 & 0.02 & 0.01 \\
    BEN & GEO+CLM vs CLM & 100\% & 99.61 & 0.28 & 0.11 \\
    BEN & COMP+CLM vs CLM & 20\% & 100.00 & 0.00 & 0.00 \\
    BEN & COMP+CLM vs CLM & 40\% & 99.96 & 0.02 & 0.02 \\
    BEN & COMP+CLM vs CLM & 60\% & 99.80 & 0.13 & 0.07 \\
    BEN & COMP+CLM vs CLM & 80\% & 98.06 & 1.19 & 0.75 \\
    BEN & COMP+CLM vs CLM & 100\% & 99.14 & 0.48 & 0.38 \\
    BEN & GEO+CLM vs COMP+CLM & 20\% & 88.98 & 1.08 & 9.94 \\
    BEN & GEO+CLM vs COMP+CLM & 40\% & 99.54 & 0.31 & 0.15 \\
    BEN & GEO+CLM vs COMP+CLM & 60\% & 58.85 & 1.47 & 39.68 \\
    BEN & GEO+CLM vs COMP+CLM & 80\% & 97.27 & 1.44 & 1.30 \\
    BEN & GEO+CLM vs COMP+CLM & 100\% & 99.79 & 0.09 & 0.12 \\
    BEN & CLM vs BASE & 20\% & 99.86 & 0.09 & 0.05 \\
    BEN & CLM vs BASE & 40\% & 0.34 & 97.66 & 1.99 \\
    BEN & CLM vs BASE & 60\% & 99.99 & 0.01 & 0.00 \\
    BEN & CLM vs BASE & 80\% & 99.96 & 0.03 & 0.01 \\
    BEN & CLM vs BASE & 100\% & 16.48 & 29.68 & 53.84 \\
    DFC & GEO+CLM vs CLM & 20\% & 97.68 & 0.45 & 1.87 \\
    DFC & GEO+CLM vs CLM & 40\% & 1.29 & 1.40 & 97.31 \\
    DFC & GEO+CLM vs CLM & 60\% & 0.24 & 1.68 & 98.08 \\
    DFC & GEO+CLM vs CLM & 80\% & 0.56 & 1.42 & 98.02 \\
    DFC & GEO+CLM vs CLM & 100\% & 0.15 & 0.67 & 99.19 \\
    DFC & COMP+CLM vs CLM & 20\% & 98.89 & 0.30 & 0.81 \\
    DFC & COMP+CLM vs CLM & 40\% & 5.01 & 0.98 & 94.01 \\
    DFC & COMP+CLM vs CLM & 60\% & 13.62 & 1.28 & 85.10 \\
    DFC & COMP+CLM vs CLM & 80\% & 4.83 & 0.59 & 94.57 \\
    DFC & COMP+CLM vs CLM & 100\% & 2.24 & 0.23 & 97.53 \\
    DFC & GEO+CLM vs COMP+CLM & 20\% & 0.01 & 0.16 & 99.83 \\
    DFC & GEO+CLM vs COMP+CLM & 40\% & 0.08 & 0.48 & 99.45 \\
    DFC & GEO+CLM vs COMP+CLM & 60\% & 0.31 & 63.84 & 35.85 \\
    DFC & GEO+CLM vs COMP+CLM & 80\% & 0.04 & 1.64 & 98.32 \\
    DFC & GEO+CLM vs COMP+CLM & 100\% & 0.10 & 16.67 & 83.23 \\
    DFC & CLM vs BASE & 20\% & 94.83 & 0.42 & 4.76 \\
    DFC & CLM vs BASE & 40\% & 99.39 & 0.07 & 0.54 \\
    DFC & CLM vs BASE & 60\% & 29.70 & 0.35 & 69.95 \\
    DFC & CLM vs BASE & 80\% & 21.49 & 0.29 & 78.22 \\
    DFC & CLM vs BASE & 100\% & 0.60 & 0.83 & 98.57 \\
    LCZ & GEO+CLM vs CLM & 20\% & 99.99 & 0.01 & 0.00 \\
    LCZ & GEO+CLM vs CLM & 40\% & 99.99 & 0.01 & 0.00 \\
    LCZ & GEO+CLM vs CLM & 60\% & 99.80 & 0.16 & 0.04 \\
    LCZ & GEO+CLM vs CLM & 80\% & 98.95 & 0.75 & 0.30 \\
    LCZ & GEO+CLM vs CLM & 100\% & 99.60 & 0.30 & 0.11 \\
    LCZ & COMP+CLM vs CLM & 20\% & 99.84 & 0.13 & 0.03 \\
    LCZ & COMP+CLM vs CLM & 40\% & 97.49 & 0.70 & 1.81 \\
    LCZ & COMP+CLM vs CLM & 60\% & 99.69 & 0.24 & 0.07 \\
    LCZ & COMP+CLM vs CLM & 80\% & 99.18 & 0.57 & 0.25 \\
    LCZ & COMP+CLM vs CLM & 100\% & 99.46 & 0.40 & 0.14 \\
    LCZ & GEO+CLM vs COMP+CLM & 20\% & 2.32 & 89.66 & 8.01 \\
    LCZ & GEO+CLM vs COMP+CLM & 40\% & 100.00 & 0.00 & 0.00 \\
    LCZ & GEO+CLM vs COMP+CLM & 60\% & 78.60 & 4.20 & 17.19 \\
    LCZ & GEO+CLM vs COMP+CLM & 80\% & 40.34 & 8.07 & 51.59 \\
    LCZ & GEO+CLM vs COMP+CLM & 100\% & 17.42 & 17.42 & 65.17 \\
    LCZ & CLM vs BASE & 20\% & 99.89 & 0.09 & 0.01 \\
    LCZ & CLM vs BASE & 40\% & 98.75 & 0.67 & 0.58 \\
    LCZ & CLM vs BASE & 60\% & 97.61 & 1.79 & 0.60 \\
    LCZ & CLM vs BASE & 80\% & 99.99 & 0.01 & 0.00 \\
    LCZ & CLM vs BASE & 100\% & 6.05 & 18.41 & 75.54 \\
    \bottomrule
\end{longtable}

\begin{table}[htb]
    \centering
    \caption{Bayesian Test Probability on Macro metrics with Rope of 0.5}
    \label{tab:bay_test_macro}
    \begin{tabular}{@{}ccc|ccc|ccc@{}}
Dataset & Model        & Epoch (\%) & \multicolumn{3}{c|}{MoCo}    & \multicolumn{3}{c}{MAE}      \\
        &              &            & P(left) & P(right) & P(rope) & P(left) & P(right) & P(rope) \\ \midrule
BEN     & GEO vs BASE  & 20\%  & 96.00   & 2.83     & 1.17    & 99.33   & 0.58     & 0.08    \\
BEN     & GEO vs BASE  & 40\%  & 98.32   & 1.27     & 0.40    & 99.95   & 0.04     & 0.01    \\
BEN     & GEO vs BASE  & 60\%  & 93.30   & 4.85     & 1.85    & 99.22   & 0.66     & 0.12    \\
BEN     & GEO vs BASE  & 80\%  & 98.04   & 1.24     & 0.72    & 96.10   & 3.17     & 0.74    \\
BEN     & GEO vs BASE  & 100\% & 97.53   & 1.71     & 0.76    & 95.84   & 2.30     & 1.86    \\
BEN     & COMP vs BASE & 20\%  & 92.36   & 5.16     & 2.48    & 98.16   & 1.59     & 0.25    \\
BEN     & COMP vs BASE & 40\%  & 97.86   & 1.59     & 0.55    & 99.69   & 0.23     & 0.09    \\
BEN     & COMP vs BASE & 60\%  & 92.50   & 5.33     & 2.17    & 98.70   & 1.08     & 0.22    \\
BEN     & COMP vs BASE & 80\%  & 98.65   & 0.86     & 0.49    & 96.42   & 2.89     & 0.68    \\
BEN     & COMP vs BASE & 100\% & 99.20   & 0.59     & 0.21    & 97.91   & 1.35     & 0.73    \\
BEN     & GEO vs COMP  & 20\%  & 81.90   & 4.07     & 14.03   & 74.48   & 15.82    & 9.70    \\
BEN     & GEO vs COMP  & 40\%  & 55.45   & 8.07     & 36.48   & 97.99   & 0.70     & 1.31    \\
BEN     & GEO vs COMP  & 60\%  & 47.68   & 32.93    & 19.39   & 89.70   & 2.27     & 8.03    \\
BEN     & GEO vs COMP  & 80\%  & 21.55   & 28.58    & 49.87   & 28.75   & 25.87    & 45.38   \\
BEN     & GEO vs COMP  & 100\% & 1.79    & 92.17    & 6.04    & 0.01    & 99.94    & 0.05    \\
DFC     & GEO vs BASE  & 20\%  & 81.19   & 3.33     & 15.47   & 99.95   & 0.04     & 0.01    \\
DFC     & GEO vs BASE  & 40\%  & 96.89   & 0.95     & 2.16    & 98.55   & 0.50     & 0.95    \\
DFC     & GEO vs BASE  & 60\%  & 99.94   & 0.03     & 0.03    & 92.70   & 0.54     & 6.76    \\
DFC     & GEO vs BASE  & 80\%  & 99.31   & 0.18     & 0.51    & 99.59   & 0.02     & 0.39    \\
DFC     & GEO vs BASE  & 100\% & 99.65   & 0.15     & 0.20    & 0.10    & 0.05     & 99.85   \\
DFC     & COMP vs BASE & 20\%  & 86.29   & 2.15     & 11.56   & 99.99   & 0.01     & 0.00    \\
DFC     & COMP vs BASE & 40\%  & 98.43   & 0.47     & 1.10    & 98.75   & 0.37     & 0.88    \\
DFC     & COMP vs BASE & 60\%  & 99.90   & 0.04     & 0.06    & 69.97   & 0.87     & 29.15   \\
DFC     & COMP vs BASE & 80\%  & 99.55   & 0.18     & 0.27    & 1.13    & 0.14     & 98.72   \\
DFC     & COMP vs BASE & 100\% & 99.45   & 0.19     & 0.36    & 0.42    & 0.94     & 98.64   \\
DFC     & GEO vs COMP  & 20\%  & 1.11    & 1.36     & 97.53   & 84.92   & 2.10     & 12.97   \\
DFC     & GEO vs COMP  & 40\%  & 2.42    & 2.80     & 94.78   & 1.85    & 0.33     & 97.82   \\
DFC     & GEO vs COMP  & 60\%  & 78.95   & 0.09     & 20.96   & 0.23    & 0.03     & 99.74   \\
DFC     & GEO vs COMP  & 80\%  & 0.05    & 96.56    & 3.39    & 95.93   & 0.01     & 4.06    \\
DFC     & GEO vs COMP  & 100\% & 0.34    & 0.00     & 99.66   & 0.63    & 0.15     & 99.22  
\end{tabular}
\end{table}

\begin{table}[htb]
\centering
\caption{Results over the three seed for MoCo and MAE}
\label{tab:main_results}
\begin{tabular}{@{}llc|c|c@{}}
\toprule
MODEL & Dataset & Epoch (\%) & MoCo              & MAE               \\ \midrule
BASE  & BEN     & 20\%       & 51.12/49.42/50.37 & 35.48/35.71/36.31 \\
BASE  & BEN     & 40\%       & 51.96/52.59/51.55 & 42.09/41.88/42.16 \\
BASE  & BEN     & 60\%       & 52.6/51.85/52.7   & 39.85/39.84/39.98 \\
BASE  & BEN     & 80\%       & 53.67/52.98/53.85 & 40.39/40.28/40.54 \\
BASE  & BEN     & 100\%      & 53.88/53.27/53.78 & 45.42/45.06/45.39 \\
COMP  & BEN     & 20\%       & 52.62/51.82/52.65 & 45.62/45.9/45.67  \\
COMP  & BEN     & 40\%       & 55.93/55.51/56.86 & 47.4/47.6/47.55   \\
COMP  & BEN     & 60\%       & 57.65/57.72/57.91 & 49.04/48.57/48.95 \\
COMP  & BEN     & 80\%       & 58.49/58.34/58.47 & 49.28/49.31/49.36 \\
COMP  & BEN     & 100\%      & 59.06/58.95/59.09 & 49.55/49.64/49.48 \\
GEO   & BEN     & 20\%       & 52.22/51.59/52.11 & 47.49/47.53/47.34 \\
GEO   & BEN     & 40\%       & 56.29/55.89/56.65 & 47.84/47.87/47.85 \\
GEO   & BEN     & 60\%       & 57.29/57.58/57.63 & 49.1/49.23/49.15  \\
GEO   & BEN     & 80\%       & 56.57/56.58/56.65 & 49.29/48.74/49.21 \\
GEO   & BEN     & 100\%      & 58.49/58.43/58.54 & 48.08/47.61/48.11 \\
BASE  & DFC     & 20\%       & 57.25/57.27/57.75 & 58.67/58.86/58.39 \\
BASE  & DFC     & 40\%       & 57.86/57.76/57.84 & 60.22/60.53/60.11 \\
BASE  & DFC     & 60\%       & 57.06/57.09/57.07 & 61.5/61.41/61.34  \\
BASE  & DFC     & 80\%       & 57.55/57.66/57.67 & 61.68/61.53/61.71 \\
BASE  & DFC     & 100\%      & 57.36/57.49/57.51 & 61.93/61.78/61.98 \\
COMP  & DFC     & 20\%       & 57.82/57.82/57.77 & 61.05/60.69/60.79 \\
COMP  & DFC     & 40\%       & 59.27/59.27/59.25 & 61.85/61.71/61.76 \\
COMP  & DFC     & 60\%       & 59.01/59.06/59.02 & 61.96/62.12/62.09 \\
COMP  & DFC     & 80\%       & 59.32/59.36/59.03 & 61.66/61.91/61.89 \\
COMP  & DFC     & 100\%      & 59.05/59.1/59.05  & 61.77/61.98/62.02 \\
GEO   & DFC     & 20\%       & 58.13/58.01/57.9  & 61.91/61.74/61.65 \\
GEO   & DFC     & 40\%       & 59.35/59.37/59.32 & 61.96/61.84/61.92 \\
GEO   & DFC     & 60\%       & 59.45/59.56/59.52 & 62.1/62.17/62.21  \\
GEO   & DFC     & 80\%       & 58.92/59.01/58.95 & 62.15/62.32/62.35 \\
GEO   & DFC     & 100\%      & 59.17/59.26/59.19 & 62.05/62.24/62.22 \\
BASE  & LCZ     & 20\%       & 75.78/74.78/74.74 & 48.68/48.3/47.3   \\
BASE  & LCZ     & 40\%       & 77.52/77.22/75.94 & 60.18/59.18/60.1  \\
BASE  & LCZ     & 60\%       & 78.36/77.44/78.04 & 61.52/60.8/58.46  \\
BASE  & LCZ     & 80\%       & 78.94/78.48/78.52 & 64.9/64.72/63.94  \\
BASE  & LCZ     & 100\%      & 79.36/78.44/78.54 & 70.36/69.56/69.92 \\
COMP  & LCZ     & 20\%       & 80.34/79.06/79.18 & 66.24/67.96/67.8  \\
COMP  & LCZ     & 40\%       & 82.32/81.96/82.26 & 72.1/72.14/72.5   \\
COMP  & LCZ     & 60\%       & 81.48/81.74/81.8  & 72.56/73.1/73.62  \\
COMP  & LCZ     & 80\%       & 82.3/82.1/82.5    & 73.06/73.04/74.22 \\
COMP  & LCZ     & 100\%      & 82.52/81.96/82.02 & 75.12/75.44/75.44 \\
GEO   & LCZ     & 20\%       & 82.42/81.42/82.16 & 70.52/70.72/72.06 \\
GEO   & LCZ     & 40\%       & 81.92/81.4/81.9   & 73.68/73.0/74.1   \\
GEO   & LCZ     & 60\%       & 82.16/82.3/82.2   & 74.64/74.82/74.62 \\
GEO   & LCZ     & 80\%       & 82.3/82.46/82.46  & 74.98/74.72/75.94 \\
GEO   & LCZ     & 100\%      & 82.28/82.54/82.74 & 73.14/73.56/74.1  \\ \bottomrule
\end{tabular}
\end{table}

\begin{table}[htb]
\centering

\caption{Results for Curriculum Masking (CLM) over three seeds}
\label{tab:clm_results}
\begin{tabular}{@{}lcc|c@{}}
\toprule
Model    & Dataset & Epoch (\%) & Performance             \\ \midrule
CLM      & BEN     & 20\%  & 40.67/40.7/40.99  \\
CLM      & BEN     & 40\%  & 40.86/40.94/41.05 \\
CLM      & BEN     & 60\%  & 45.11/45.24/45.31 \\
CLM      & BEN     & 80\%  & 45.26/44.92/45.19 \\
CLM      & BEN     & 100\% & 45.46/45.33/44.57 \\
COMP+CLM & BEN     & 20\%  & 46.12/46.11/46.34 \\
COMP+CLM & BEN     & 40\%  & 43.48/43.53/43.77 \\
COMP+CLM & BEN     & 60\%  & 49.77/49.76/49.44 \\
COMP+CLM & BEN     & 80\%  & 48.9/49.69/48.52  \\
COMP+CLM & BEN     & 100\% & 48.64/48.53/48.5  \\
GEO+CLM  & BEN     & 20\%  & 47.04/47.15/46.95 \\
GEO+CLM  & BEN     & 40\%  & 48.6/48.69/48.11  \\
GEO+CLM  & BEN     & 60\%  & 50.15/50.25/50.21 \\
GEO+CLM  & BEN     & 80\%  & 51.98/51.93/52.05 \\
GEO+CLM  & BEN     & 100\% & 50.98/51.1/51.1   \\
CLM      & DFC     & 20\%  & 59.62/59.56/59.31 \\
CLM      & DFC     & 40\%  & 61.25/61.57/61.04 \\
CLM      & DFC     & 60\%  & 61.91/61.96/61.73 \\
CLM      & DFC     & 80\%  & 62.15/62.01/62.05 \\
CLM      & DFC     & 100\% & 61.95/61.75/61.87 \\
COMP+CLM & DFC     & 20\%  & 60.98/61.21/60.99 \\
COMP+CLM & DFC     & 40\%  & 61.51/61.65/61.32 \\
COMP+CLM & DFC     & 60\%  & 62.27/62.1/62.14  \\
COMP+CLM & DFC     & 80\%  & 62.44/62.16/62.37 \\
COMP+CLM & DFC     & 100\% & 62.23/61.95/62.18 \\
GEO+CLM  & DFC     & 20\%  & 60.66/60.89/60.69 \\
GEO+CLM  & DFC     & 40\%  & 61.26/61.46/61.11 \\
GEO+CLM  & DFC     & 60\%  & 61.69/61.67/61.55 \\
GEO+CLM  & DFC     & 80\%  & 62.08/61.81/61.97 \\
GEO+CLM  & DFC     & 100\% & 61.81/61.53/61.68 \\
CLM      & LCZ     & 20\%  & 62.4/61.44/61.8   \\
CLM      & LCZ     & 40\%  & 63.08/62.84/62.92 \\
CLM      & LCZ     & 60\%  & 67.12/66.42/66.64 \\
CLM      & LCZ     & 80\%  & 70.72/70.64/69.72 \\
CLM      & LCZ     & 100\% & 70.04/69.7/69.52  \\
COMP+CLM & LCZ     & 20\%  & 71.44/71.46/70.9  \\
COMP+CLM & LCZ     & 40\%  & 64.62/64.14/64.76 \\
COMP+CLM & LCZ     & 60\%  & 74.1/74.56/74.6   \\
COMP+CLM & LCZ     & 80\%  & 75.44/76.18/75.84 \\
COMP+CLM & LCZ     & 100\% & 75.76/76.66/76.38 \\
GEO+CLM  & LCZ     & 20\%  & 70.5/69.78/69.92  \\
GEO+CLM  & LCZ     & 40\%  & 74.46/74.0/74.48  \\
GEO+CLM  & LCZ     & 60\%  & 75.26/74.94/75.88 \\
GEO+CLM  & LCZ     & 80\%  & 76.0/76.06/76.56  \\
GEO+CLM  & LCZ     & 100\% & 76.0/76.14/76.66  \\ \bottomrule
\end{tabular}
\end{table}

\begin{table}[htb]
\centering
\caption{Results for MAE and MoCo on Macro Metrics over three seeds}
\label{tab:macro_results}
\begin{tabular}{@{}lcc|c|c@{}}
\toprule
Model & Dataset & Epoch & MoCo              & MAE               \\ \midrule
BASE  & BEN     & 20\%  & 40.95/37.69/38.12 & 16.2/17.88/17.09  \\
BASE  & BEN     & 40\%  & 41.31/41.67/40.98 & 26.08/25.78/25.73 \\
BASE  & BEN     & 60\%  & 44.6/43.14/45.38  & 23.68/21.57/20.59 \\
BASE  & BEN     & 80\%  & 45.67/43.93/45.7  & 26.91/22.77/22.78 \\
BASE  & BEN     & 100\% & 44.92/43.87/45.29 & 27.53/29.72/28.77 \\
COMP  & BEN     & 20\%  & 43.07/42.81/43.43 & 30.01/27.76/32.06 \\
COMP  & BEN     & 40\%  & 47.35/46.85/48.85 & 31.44/31.82/32.72 \\
COMP  & BEN     & 60\%  & 48.32/50.32/48.88 & 32.36/32.63/32.19 \\
COMP  & BEN     & 80\%  & 49.26/48.91/49.97 & 33.31/32.12/32.66 \\
COMP  & BEN     & 100\% & 50.81/51.17/51.16 & 33.38/32.92/32.94 \\
GEO   & BEN     & 20\%  & 44.48/43.27/44.72 & 30.85/32.14/32.83 \\
GEO   & BEN     & 40\%  & 48.55/47.16/49.07 & 33.47/33.88/34.23 \\
GEO   & BEN     & 60\%  & 50.89/49.64/48.15 & 33.83/33.35/33.56 \\
GEO   & BEN     & 80\%  & 49.72/49.0/49.14  & 32.74/33.01/32.46 \\
GEO   & BEN     & 100\% & 49.18/50.29/49.89 & 32.1/31.6/31.6    \\
BASE  & DFC     & 20\%  & 41.18/41.43/41.29 & 36.43/37.07/36.18 \\
BASE  & DFC     & 40\%  & 41.6/40.96/41.49  & 42.72/43.43/42.36 \\
BASE  & DFC     & 60\%  & 40.74/40.83/41.01 & 44.32/44.44/44.05 \\
BASE  & DFC     & 80\%  & 41.16/41.41/41.55 & 44.47/44.21/44.33 \\
BASE  & DFC     & 100\% & 40.97/41.27/41.43 & 44.91/44.53/44.72 \\
COMP  & DFC     & 20\%  & 42.49/42.05/42.27 & 44.13/43.04/43.36 \\
COMP  & DFC     & 40\%  & 43.06/42.88/43.12 & 44.88/44.22/44.39 \\
COMP  & DFC     & 60\%  & 43.0/43.16/43.17  & 44.87/44.8/44.91  \\
COMP  & DFC     & 80\%  & 43.46/43.64/43.52 & 44.42/44.55/44.76 \\
COMP  & DFC     & 100\% & 43.03/43.24/43.17 & 44.46/44.55/44.85 \\
GEO   & DFC     & 20\%  & 42.56/41.95/42.22 & 44.89/44.33/44.05 \\
GEO   & DFC     & 40\%  & 43.04/42.99/42.97 & 45.05/44.5/44.56  \\
GEO   & DFC     & 60\%  & 43.49/43.72/43.75 & 45.12/45.03/45.12 \\
GEO   & DFC     & 80\%  & 42.79/43.0/42.93  & 44.98/45.13/45.3  \\
GEO   & DFC     & 100\% & 43.47/43.68/43.6  & 44.61/44.77/45.0  \\ \bottomrule
\end{tabular}
\end{table}

\end{document}